\documentclass[11pt]{article}

\usepackage{acl2015}
\usepackage{times}
\usepackage{url}
\usepackage{latexsym}

\usepackage{graphicx}
\usepackage{color}
\usepackage{booktabs}
\usepackage{subcaption}
\usepackage{multirow}
\usepackage{amsmath}
\usepackage{amssymb}

\DeclareMathOperator \real{\mathbb{R}}

\DeclareMathOperator*{\Ber}{Ber}
\DeclareMathOperator*{\sig}{sig}

\newcommand{\disco}{\nobreak{DiscoTK}}

\newcommand{\spede}{\nobreak{\sc spede07pP}}

\newcommand{\amber}{\nobreak{AMBER}}
\newcommand{\meteor}{\nobreak{\sc Meteor}}

\newcommand{\simpbleu}{\nobreak{SIMPBLEU}}
\newcommand{\ter}{\nobreak{TER}}
\newcommand{\bleu}{\nobreak{BLEU}}

\newcommand{\nist}{\nobreak{NIST}}

\newcommand{\Ni}{({\em i})~}
\newcommand{\Nii}{({\em ii})~}
\newcommand{\Niii}{({\em iii})~}

\newcommand{\metrics}{\nobreak{\sc 4metrics}}
\newcommand{\syntax}{\nobreak{\sc syntax25}}
\newcommand{\wiki}{\nobreak{\sc Wiki-GW25}}
\newcommand{\wikii}{\nobreak{\sc Wiki-GW300}}
\newcommand{\commoncrawl}{\nobreak{\sc CC-300-42B}}
\newcommand{\commoncrawll}{\nobreak{\sc CC-300-840B}}
\newcommand{\composes}{\nobreak{\sc Composes400}}
\newcommand{\wordvec}{\nobreak{\sc word2vec300}}
\newcommand{\bleucomp}{\nobreak{\sc BLEUcomp}}

\title{Pairwise Neural Machine Translation Evaluation}

\author{Francisco Guzm\'an\hspace*{3mm}
Shafiq Joty\hspace*{3mm}
Llu\'is M\`arquez\hspace*{2mm}\hbox{\rm and}\hspace*{2mm}Preslav Nakov\\
ALT Research Group\\
Qatar Computing Research Institute --- HBKU, Qatar Foundation\\
{\tt\{fguzman,sjoty,lmarquez,pnakov\}@qf.org.qa}}

\date{}

\begin{document}
 
%----------------------- T i t l e
\maketitle
%\footnotetext{Draft. To appear in ACL 2015}
%----------------------- A b s t r a c t
\begin{abstract}
We present a novel framework for machine translation evaluation using neural networks in a pairwise setting, where the goal 
is to select the better translation from a pair of hypotheses, given the reference 
translation. In this framework, lexical, syntactic and semantic information from the 
reference and the two hypotheses is compacted into relatively small distributed vector 
representations, and fed into a multi-layer neural network that models the interaction between 
each of the hypotheses and the reference, as well as between the two hypotheses.
These compact representations are in turn based on word and sentence embeddings,
which are learned using neural networks.
The framework is flexible, allows for efficient learning and classification,
and yields correlation with humans that rivals the state of the art. 
\end{abstract}

%----------------------- S e c t i o n s
\section{Introduction}
\label{sec:intro}
%-------------------------------------------------------
% Neural Machine Translation Evaluation
% Paper submitted to NAACL-2105
% December 2014
%-------------------------------------------------------
% Section 1: Introduction
%-------------------------------------------------------

Automatic machine translation (MT) evaluation is a necessary step when developing or comparing MT systems. \emph{Reference}-based MT evaluation, i.e., comparing the system output to one or more human reference translations, is the most common approach. Existing MT evaluation measures typically output an absolute quality score by computing the similarity between the machine and the human translations.
%, and the source sentence is  usually ignored in this framework.
%
In the simplest case, the similarity is computed by counting word $n$-gram matches between the translation and the reference. This is the case of \bleu~\cite{Papineni:Roukos:Ward:Zhu:2002}, which has been the standard for MT evaluation for years. Nonetheless, more recent evaluation measures take into account various aspects of linguistic similarity, and achieve better correlation with human judgments. 
%for instance, synonymy and paraphrasing~\cite{Lavie:2009:MMA}, syntax~\cite{Gimenez2007,Popovic2007,Liu2005}, semantics~\cite{Gimenez2007,Lo2012}, and discourse~\cite{Comelles2010,Wong2012,discoMT:acl2014,discoMT:WMT2014}. The combination of all these aspects led to improved results in metric evaluation campaigns, such as the \emph{WMT metrics task}~\cite{WMT14}. 

Having absolute quality scores at the sentence level allows to rank alternative translations for a given source sentence. This is useful, for instance, for statistical machine translation (SMT) parameter tuning, for system comparison, and for assessing the progress during MT system development. The quality of automatic MT evaluation metrics is usually assessed by computing their correlation with human judgments. To that end, quality rankings of alternative translations have been created by human judges. It is known that assigning an absolute score to a translation is a difficult task for humans. Hence, ranking-based evaluations, where judges are asked to rank the output of \hbox{2 to 5} systems, have been used in recent years, which has yielded much higher inter-annotator agreement \cite{callisonburch-EtAl:2007:WMT}.

These human quality judgments can be used to train automatic metrics.
% that reproduce human annotations.
This supervised learning can be oriented to predict absolute scores, e.g., using regression~\cite{albrecht:2008}, or rankings~\cite{Duh:2008,song-cohn:2011:WMT}. A particular case of the latter is used to learn in a pairwise setting, i.e., given a reference and two alternative translations (or hypotheses), the task is to decide which one is better. This setting emulates closely how human judges perform evaluation assessments in reality, and can be used to produce rankings for an arbitrarily large number of hypotheses.
In this pairwise setting, the challenge is to learn, from a pair of hypotheses, which are the features that help to discriminate the better from the worse translation. 
Although the pairwise setting does not produce absolute quality scores (i.e.,~it is not an evaluation metric applicable to a single translation), it is useful and arguably sufficient for most evaluation and MT development scenarios.\footnote{We do not argue that the pairwise approach is better than the direct estimation of human quality scores. Both approaches have pros and cons; we see them as complementary.}

Recently, Guzm\'an et al.~\shortcite{guzman-EtAl:2014} presented a learning framework for this pairwise setting, based on preference kernels and support vector machines (SVM). They obtained promising results using syntactic and discourse-based structures. However, using convolution kernels over complex structures comes at %the price of 
a high computational cost %. This inefficiency applies 
both at training and at testing time because the use of kernels requires that the SVM operate in the much slower dual space. Thus, some simplification is needed to make it practical.
While there are some solutions in the kernel-based learning framework to alleviate the computational burden, in this paper we explore an entirely different direction.
%Apart from addressing the aforementioned learning challenge authors claimed that their framework is flexible and allows the addition of different layers of linguistic information in the form of tree-like structures. Results presented were promising in both directions. The main disadvantage of the kernel approach is that it does not scale well to large datasets and complex relational structures (graphs or enriched tres). This inefficiency applies bot at training and test time. Thus, some simplifications are needed to make it practical. In the kernel-based learning framework there are some solutions to alleviate the computational problem, but in this paper we explore a totally different direction.

We present a novel neural-based architecture for learning in the pairwise setting for MT evaluation. Lexical, syntactic and semantic information from the reference and the two hypotheses is compacted into relatively small distributed vector representations and fed into the input layer, together with a set of individual real-valued features coming from simple pre-existing MT evaluation metrics. 
A hidden layer, motivated by our intuitions on the pairwise ranking problem, is used to capture interactions between the relevant input components. Finally, we present a task-oriented cost function, specifically tailored for this problem.

Our evaluation results on the \emph{WMT12 metrics task} benchmark datasets~\cite{WMT12} show very high correlation with human judgments.
These results clearly surpass~\cite{guzman-EtAl:2014} and are comparable to the best previously reported results for this dataset, achieved by DiscoTK~\cite{discoMT:WMT2014}, which is a much heavier combination-based metric. 

%Apart from that, 
Another advantage of the proposed architecture is efficiency. Due to the vector-based compression of the linguistic structure and the relatively reduced size of the network, %learning and, especially, 
testing is fast, which would greatly facilitate the practical use of this approach in real MT evaluation and development. Finally, we empirically show that syntactically- and semantically-oriented embeddings can be incorporated to produce sizeable and cumulative gains in performance over a strong combination of pre-existing MT evaluation measures (\bleu, \nist, \meteor, and \ter). This is promising evidence towards our longer-term goal of defining a general platform for integrating varied linguistic information and for producing more informed MT evaluation measures.

%The contribution of the paper is twofold: i) using distributed representations of syntax and lexical semantics for MT evaluation in a flexible combination framework; ii) training a NN architecture with a hidden layer to capture interesting pairwise interactions. We don't consider the distributed representations of syntax and semantics a minor aspect, as they capture information not present in a strong combination of individual lexical evaluation metrics (BLEU+NIST+METEOR+TER). Kendall's Tau goes from 27.06 to 29.70, a sizeable improvement according to the state-of-the-art. 

\section{Related Work}
\label{sec:related}
%!TEX root = neural-mte.tex
%-------------------------------------------------------
% Neural Machine Translation Evaluation
% Paper submitted to NAACL-2105
% December 2014
%-------------------------------------------------------
% Section 2: Related work
%-------------------------------------------------------

Contemporary MT evaluation measures have evolved beyond %\bleu \ and 
simple lexical matching, and now take into account various aspects of linguistic structures,
including synonymy and paraphrasing~\cite{Lavie:2009:MMA}, syntax~\cite{Gimenez2007,Popovic2007,Liu2005}, semantics~\cite{Gimenez2007,Lo2012}, and even discourse~\cite{Comelles2010,Wong2012,discoMT:acl2014,discoMT:WMT2014}. The combination of several of these aspects has led to improved results in metric evaluation campaigns, such as the \emph{WMT metrics task}~\cite{WMT14}.
 
In this paper, we present a general framework for learning  to rank translations in the pairwise setting, using information from several linguistic representations of the translations and references.  This work has connections with the ranking-based approaches for learning to reproduce human judgments of MT quality.
In particular, our setting is similar to that of \newcite{Duh:2008}, but differs from it  %we extend it to a new level, 
both in terms of the feature 
representation and of the learning framework. For instance, we integrate several layers of linguistic information, while \newcite{Duh:2008} only used 
lexical and POS matches as features. Secondly, we use information about both the reference and the two alternative translations simultaneously %
%Finally, %instead of deciding upfront which types of features between hypotheses and references are important, 
%we 
in a neural-based learning framework %
capable of modeling complex interactions between the features. %framework to generate and select them automatically. 

Another related work is that of \newcite{kulesza2004}, in which lexical and syntactic features, together with other metrics, e.g., BLEU and NIST, are used in an SVM classifier to discriminate good from bad translations.
%Although related, there are important differences between the two approaches.
However, their setting is not pairwise comparison, but a classification task to distinguish \emph{human}- from \emph{machine-produced} translations. Moreover, in their work, using syntactic features decreased the correlation with human judgments dramatically (although classification accuracy improved), while in our case the effect is positive.

In our previous work \cite{guzman-EtAl:2014}, we introduced a learning framework for the pairwise setting, based on preference kernels and SVMs. We used lexical, POS, syntactic and discourse-based information in the form of tree-like structures to learn to differentiate better from worse translations.

However, in that work we used convolution kernels, which is computationally expensive and does not scale well to large datasets and complex structures such as graphs and enriched trees. This inefficiency arises both at training and testing time. 
%Unlike that work,
%here we use a different learning approach that makes
Thus, here
we use neural embeddings and multi-layer neural networks,
which yields an efficient learning framework that works significantly better on the same datasets
(although we are not using exactly the same information for learning).

%\comm{The following needs to be developed but the ideas are there. Please, Shafiq, go ahead and complete this part of the related work. Try to be brief.}

%\noindent To the best of our knowledge, the application of structured neural embeddings and neural network learning for MT evaluation is completely novel. 
%But of course, a lot of effort has been devoted in the recent years on deep NNs and embeddings for NLP. As a result, some resources and tools are already available to the community. \comm{Some selected references needed here.}
%Therefore, it is related to the aforementioned research line. 

To the best of our knowledge, the application of structured neural embeddings and a neural network learning architecture for MT evaluation is completely novel. This is despite the growing interest in recent years for deep neural nets (NNs) and word embeddings with application to a myriad of NLP problems. For example, in SMT we have observed an increased use of neural nets for language modeling ~\cite{Bengio03,Mikolov10} as well as for improving the translation model \cite{Devlin14,SutskeverVL14}. 
 %specifically by using %. The approaches vary from combining language models (LMs) with word embeddings using NNs to produce 
%neural network language models (NNLMs)\cite{Bengio03}, %to the more sophisticated 
%and recurrent neural networks (RNNs)\cite{Mikolov10}. %further improved on this by using . 
%NNLMs and RNNs have been recently used to improve SMT \cite{Devlin14,kalchbrenner13,SutskeverVL14}.

Deep learning %using NNs 
has spread beyond language modeling. For example, recursive NNs have been used for syntactic parsing \cite{socher-EtAl:2013:ACL2013} and sentiment analysis \cite{socher-EtAl:2013:EMNLP}. %, where they build a fixed-length vector representation (i.e., embedding) for a sentence using the parse tree as a backbone and composing the embeddings of its parts. 
The increased use of NNs by the NLP community is in part due to \Ni the emergence of tools such as word2vec \cite{MikolovNIPS2013} and  GloVe \cite{pennington-socher-manning:2014:EMNLP2014}, which have enabled NLP researchers to learn word embeddings, % based on skip-grams, and has been enabling for many NLP researchers. Also, %We should also mention the work of 
and \Nii unified learning frameworks, e.g., \cite{Collobert11}, which cover a variety of NLP tasks such as part-of-speech tagging, chunking, named entity recognition, and semantic role labeling. %  has proposed a unified NN framework applicable to various NLP tasks . 

While in this work we make use of %instantiate the linguistic layers with 
%two types of 
widely available pre-computed structured embeddings, %(%applied 
%at the sentence level: 
%a semantic word-based representation and a dependency-parse syntactic representation).
% Yet, 
the novelty of our work goes beyond the type of information considered as input, and resides on the way it is integrated to a neural network architecture that is inspired by our intuitions about MT evaluation. % the form of embeddings.

%%% Preslav: I have just removed this. Which tool we use is not interesting, and the new function is not central.
%We are not inventing anything new here: 1) In this work we take advantage of Socher's NN parser and several precalculated distributed representations of words; 2) Also, we use Theano and standard NN technology and methods to implement and train our learners. Our main contribution on this ML side is the implementation of a cost function that is specifically tailored for the MT evaluation task, as it optimizes directly the task evaluation measure (Kendall's $\tau$).  \comm{References needed in this paragraph.}

\section{Neural Ranking Model}
%\section{Neural architecture for MT evaluation}
\label{sec:architecture}
%!TEX root = neural-mte.tex
%-------------------------------------------------------
% Neural Machine Translation Evaluation
% Paper submitted to NAACL-2105
% December 2014
%-------------------------------------------------------
% Section 3: Neural architecture for MT evaluation
%-------------------------------------------------------

%\begin{itemize}

%\item Define the learning setting and objects 
%Besides their growing popularity,
Our motivation for using neural networks for MT evaluation is twofold. First, to take advantage of their ability to model complex non-linear relationships efficiently. Second, to have a framework that allows for easy incorporation of rich syntactic and semantic representations captured by word embeddings, which are in turn learned using deep learning.
%Below, we describe the learning task, and the neural network architecture we propose for it.

\subsection{Learning Task}
Given two translation hypotheses $t_1$ and $t_2$ (and a reference translation $r$), we want to tell which of the two is better.\footnote{In this work, we do not learn to predict ties, and ties are excluded from our training data.} Thus, we have a binary classification task, which is modeled by the class variable $y$, defined as follows:
\begin{equation}
y = \left\{ 
\begin{array}{c c}
  1 & \mbox{if $t_1$ is better than $t_2$ given $r$}\\
  0 &  \mbox{if $t_1$ is worse than $t_2$ given $r$}\\ \end{array} \right. 
\end{equation}

We model this task using a feed-forward neural network (NN) of the form:
\begin{equation}
p(y|t_1,t_2,r)= \Ber(y|f(t_1,t_2,r))
\end{equation}

\noindent which is a Bernoulli distribution of $y$ with parameter $\sigma = f(t_1,t_2,r)$,
defined as follows:
\begin{equation}
f(t_1,t_2,r)=\sig(\mathbf{w^T_v}\phi(t_1,t_2,r) + b_v) \label{outfunc}
\end{equation}

\noindent where $\sig$ is the sigmoid function, $\phi(x)$ defines the transformations of the input $x$ through the hidden layer, $\mathbf{w_v}$ are the weights from the hidden layer to the output layer, and $b_v$ is a bias term. 

\subsection{Network Architecture}

\begin{figure}[ht]
\centering
\includegraphics[width=.5\textwidth]{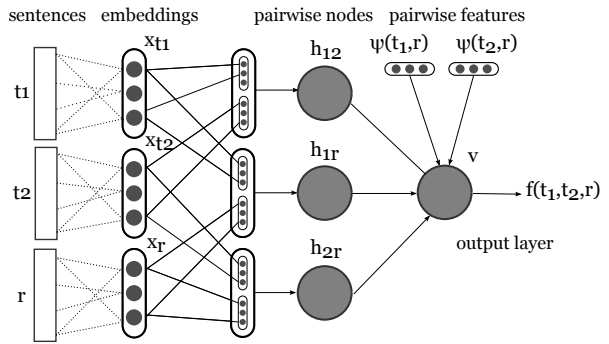}
\caption{\label{fig:architecture} {\small Overall architecture of the neural network.}}
\end{figure}

In order to decide which hypothesis is \emph{better} given the tuple $(t_1,t_2,r)$ as input, we first map the hypotheses and the reference to a fixed-length vector $\left[ \mathbf{x}_{t_1},  \mathbf{x}_{t_2}, \mathbf{x}_r \right]$, using syntactic and semantic embeddings. Then, we feed this vector as input to our neural network, whose architecture is shown in Figure~\ref{fig:architecture}. 

%\item Motivate and describe the learning architecture (in general). Include a figure with a graphical representation.
In our architecture, we model three types of interactions, using different groups of nodes in the hidden layer. 
We have two \emph{evaluation} groups $\mathbf{h_{1r}}$ and $\mathbf{h_{2r}}$
% which are inspired by traditional machine translation evaluation metrics
that model how similar each hypothesis $t_i$ is to the reference $r$.

The vector representations of the hypothesis (i.e.,~$\mathbf{x}_{t1}$ or $\mathbf{x}_{t2}$) together with the reference (i.e.,~$\mathbf{x}_{r}$) constitute the input to the hidden nodes in these two groups. The third group of hidden nodes $\mathbf{h_{12}}$, which we call \emph{similarity} group, models how close $t_1$ and $t_2$ are.
This might be useful as 
%, in order to differentiate a better translation from a worse translation, our model should learn weights not only by comparing the candidate translations with the reference but also by comparing the two hypotheses as
highly similar hypotheses are likely to be comparable in quality, irrespective of whether they are good or bad in absolute terms.

The input to each of these groups is represented by concatenating the vector representations of the two components participating in the interaction, i.e., $\mathbf{x_{1r}} = \left[ \mathbf{x}_{t_1}, \mathbf{x}_r \right]$, $\mathbf{x_{2r}} = \left[\mathbf{x}_{t_2}, \mathbf{x}_{r} \right]$, $\mathbf{x_{12}} = \left[ \mathbf{x}_{t_1}, \mathbf{x}_{t_2} \right]$. In summary, the transformation $\phi(t_1,t_2,r) = [\mathbf{h_{12}}, \mathbf{h_{1r}}, \mathbf{h_{2r}}]$ in our NN architecture can be written as follows:   
\begin{eqnarray*}
 \mathbf{h_{1r}} &=& g( \mathbf{W_{1r}} \mathbf{x_{1r}} + \mathbf{b_{1r}})\\
 \mathbf{h_{2r}} &=& g( \mathbf{W_{2r}} \mathbf{x_{2r}} +  \mathbf{b_{2r}})\\
 \mathbf{h_{12}} &=& g( \mathbf{W_{12}} \mathbf{x_{12}} +  \mathbf{b_{12}})
\end{eqnarray*}
%\vspace{-2pt}
\noindent where $g(.)$ is a non-linear activation function (applied component-wise), $\mathbf{W} \in \real^{H \times N}$ are the associated weights between the input layer and the hidden layer, and $\mathbf{b}$ are the corresponding bias terms. In our experiments, we used $\tanh$ as an activation function, rather than $\sig$, to be consistent with how parts of our input vectors were generated.\footnote{Many of our input representations consist of word embeddings trained with neural networks that used $\tanh$ as an activation function.} 
%
%\noindent where $x_{12},x_{1r},x_{2r} \in \real^{2N}$ are our pairwise input vectors, $\mathbf{W_{1r}} \in \real^{H \times 2N}$ and $\mathbf{W_{2r}}\in \real^{H \times 2N}$ are the weights for the evaluation layers each with $H$ hidden units, $\mathbf{W_{12}}\in \real^{H \times 2N}$ are the weights for the similarity layer,$\mathbf{W}^T_v \in \real^{1 \times 3H+2\Phi}$ are the weights for the output layer, and $\Phi$ is the dimensionality of the pairwise feature-vectors $\mathbf{\phi}$. Finally, $\tanh(\cdot)$ and $\sig(\cdot)$ are the element-wise hyperbolic tangent and sigmoid functions, respectively.

In addition, our model allows to incorporate external sources of information by enabling \emph{skip arcs} that go directly from the input to the output, skipping the hidden layer. In our setting, these arcs represent pairwise similarity features between the translation hypotheses and the reference (e.g., the \bleu~ scores of the translations). We denote these pairwise external feature sets as $\mathbf{\psi}_{1r}=\psi(t_1,r)$ and  $\mathbf{\psi}_{2r}=\psi(t_2,r)$. When we include the external features in our architecture, the activation at the output, i.e., eq.~(\ref{outfunc}), can be rewritten as follows:  
\begin{equation*}
f(t_1,t_2,r)=\sig(\mathbf{w^T_v} [\phi(t_1,t_2,r), \mathbf{\psi}_{1r}, \mathbf{\psi}_{2r}] + b_v) 
\end{equation*}

\subsection{Network Training}
\label{logcost}
The negative log likelihood of the training data for the model parameters \mbox{$\theta=(\mathbf{W_{12}}, \mathbf{W_{1r}}, \mathbf{W_{2r}, w_v}, \mathbf{b_{12}}, \mathbf{b_{1r}}, \mathbf{b_{2r}}, b_v)$}
can be written as follows:
\begin{equation}
J_{\mathbf{\theta}} = \\
- \sum_{n} y_n \log \hat{y}_{n\theta} + (1-y_n) \log \left(1- \hat{y}_{n\theta} \right) 
\end{equation}

In the above formula, $\hat{y}_{n\theta}=f_n(t_1,t_2,r)$ is the activation at the output layer for the $n$-th data instance. It is also common to use a regularized cost function by adding a weight decay penalty (e.g.,~$L_2$ or $L_1$ regularization) and to perform maximum aposteriori (MAP) estimation of the parameters. We trained our network with stochastic gradient descent (SGD), mini-batches and adagrad updates~\cite{Duchi11}, using Theano~\cite{bergstra+al:2010-scipy}.

\section{Experimental Setup}
\label{sec:setup}
%-------------------------------------------------------
% Neural Machine Translation Evaluation
% Paper submitted to NAACL-2105
% December 2014
%-------------------------------------------------------
% Section 4: Experimental setup
%-------------------------------------------------------

%\begin{itemize}
%\item Subsection 1: WMT corpora description and task evaluation measures
%\item Subsection 2: Explain the training setting for the NN model: early stopping, parameters and parameter tuning, pre-training, etc.
%\end{itemize}

In this section, we describe the different aspects of our general experimental setup (we will discuss some extensions thereof in Section~\ref{sec:discussion}), starting with a description of the input representations we use to capture the syntactic and semantic characteristics of the two hypothesis translations and the corresponding reference, as well as the datasets used to evaluate the performance of our model.

\subsection{Word Embedding Vectors}

Word embeddings play a crucial role in our model, since they allow us to model complex relations between the translations and the reference using syntactic and semantic vector representations. \\
 %In our experiments, we experiment with syntactic and semantic vectors as a representation for a sentence.

\vspace{-2mm}
\noindent{\bf Syntactic vectors}. 
We generate a syntactic vector for each sentence\
using the Stanford neural parser \cite{socher-EtAl:2013:ACL2013},
which generates a 25-dimensional vector as a by-product of syntactic parsing using a recursive NN.
Below we will refer to these vectors as \syntax.\\

\vspace{-2mm}
\noindent{\bf Semantic vectors}. 
We compose a semantic vector for a given sentence
using the average of the embedding vectors for the words it contains~\cite{Mitchell:Lapata:2010}.
We use pre-trained, fixed-length word embedding vectors produced by 
\Ni GloVe \cite{pennington-socher-manning:2014:EMNLP2014},
\Nii COMPOSES \cite{P14-1023},
and
\Niii word2vec \cite{mikolov-yih-zweig:2013:NAACL-HLT}.

Our primary representation is based on 50-dimensional GloVe vectors,
trained on Wikipedia 2014+Gigaword 5 (6B tokens),
to which below we will refer as \wiki. 

Furthermore, we experiment with \wikii, the 300-dimensional GloVe vectors trained on the same data,
as well as with the \commoncrawl \ and \commoncrawll,
300-dimensional GloVe vectors trained on 42B and on 840B tokens from Common Crawl.

We also experiment with the pre-trained, 300-dimensional word2vec embedding vectors, or \wordvec,
trained on 100B words from Google News.
Finally, we use \composes, the 400-dimensional COMPOSES vectors trained on 2.8 billion tokens
from ukWaC, the English Wikipedia, and the British National Corpus.

\subsection{Tuning and Evaluation Datasets}

We experiment with datasets of segment-level human rankings of system outputs
from the WMT11, WMT12 and WMT13 Metrics shared tasks \cite{WMT11,WMT12,WMT13}.
We focus on translating into English,
for which the WMT11 and WMT12 datasets can be split by source language:
Czech (cs), German (de), Spanish (es), and French (fr);
WMT13 also has Russian (ru).
%There were about 10,000 non-tied human judgments per language pair per dataset.

\subsection{Evaluation Score}

We evaluate our metrics in terms of correlation with human judgments measured using Kendall's $\tau$.
We report $\tau$ for the individual languages as well as macro-averaged across all languages.

Note that there were different versions of $\tau$ at WMT over the years.
Prior to 2013, WMT used a strict version, which was later relaxed at WMT13 and further revised at WMT14.
See \cite{machacek-bojar:2014:W14-33} for a discussion.
Here we use the strict version used at WMT11 and WMT12.

\subsection{Experimental Settings}

{\bf Datasets}: We train our neural models on WMT11 and we evaluate them on WMT12. 
We further use a random subset of 5,000 examples from WMT13 as a validation set to implement early stopping.

\noindent {\bf Early stopping}: We train on WMT11 for up to 10,000 epochs, and we calculate Kendall's $\tau$ on the development set after each epoch.
We then select the model that achieves the highest $\tau$ on the validation set; in case of ties for the best $\tau$, we select the latest epoch that achieved the highest $\tau$. 

\noindent{\bf Network parameters}: We train our neural network using SGD with adagrad, an initial learning rate of $\eta=0.01$, mini-batches of size $30$, and $L_2$ regularization with a decay parameter $\lambda =1e^{-4}$. We initialize the weights for our matrices by sampling from a uniform distribution following \cite{Xavier10}. We further set the size of each of our pairwise hidden layers $H$ to four nodes, and we normalize the input data using min-max to map the feature values to the range $[-1,1]$.

\section{Experiments and Results}
\label{sec:results}
%-------------------------------------------------------
% Neural Machine Translation Evaluation
% Paper submitted to NAACL-2105
% December 2014
%-------------------------------------------------------
% Section 5: Results
%-------------------------------------------------------

\begin{table*}[tb!h]
\centering
\vspace{-7mm}
{\small\begin{tabular}{cllccccc}
\toprule
& \multicolumn{1}{c}{\bf System}& \multicolumn{1}{c}{\bf Details} & \multicolumn{5}{c}{\bf Kendall's $\tau$} \\\cmidrule(l{2pt}r{2pt}){4-8}

{\bf I}&\multicolumn{2}{l}{\bf \metrics: commonly-used individual metrics}& { \bf cz }&{\bf de} & {\bf es}  & {\bf fr} &{\bf AVG}\\
\midrule

&\bleu & no learning & 15.88 & 18.56 & 18.57 & 20.83 & 18.46\\
&\nist & no learning & 19.66 & 23.09 & 20.41 & 22.21 & 21.34\\
&\ter & no learning & 17.80 & 25.31& 22.86 & 21.05 & 21.75\\
&\meteor & no learning & 20.82 &26.79 & 23.81 & 22.93 & 23.59\\\vspace{-2mm}
\\

{\bf II}&\multicolumn{2}{l}{\bf NN using embedding vectors: syntactic \& semantic}\\
\midrule
&\syntax & multi-layer NN & 8.00 & 13.03 & 12.11 & 7.42 & 10.14 \\
&\wiki & multi-layer NN & 14.31 & 11.49 & 9.24 & 4.99 & 10.01 \\\vspace{-2mm}
%\hline
%BLEU-comp & logistic regression & 18.18 & 21.13 & 19.79 & 19.91 & 19.75\\
%BLEU-comp+syntax25 & NN & 20.75 & 25.32 & 24.85 & 23.88 & 23.70\\
%BLEU-comp+wiki50 & NN & ???\\
%BLEU-comp+syntax25+wiki25 & NN & 22.42 & 28.00 & 27.24 & 24.99 & 25.66\\
\\
{\bf III}&\multicolumn{2}{l}{\bf NN using \metrics + embedding vectors}\\
\midrule
&\metrics &  logistic regression & 23.46 & 29.95 & 27.49 & 27.36 & 27.06\\
&\metrics+\syntax & multi-layer NN & 26.09 & 30.58 & 29.30 & 28.07 & 28.51 \\
&\metrics+\wiki & multi-layer NN & 25.67 & 32.50 & 29.21 & 28.92 & 29.07 \\
&\metrics+\syntax+\wiki & multi-layer NN & 26.30 & 33.19 & 30.38& 28.92 & {\bf 29.70} \\\vspace{-2mm}
\\
%LEXICAL+syntax25+wiki50(PT) & NN with pretraining & & & & & 29.48 \\
%\hline 
%LEXICAL+syntax25+wiki50 & NN no $h_1$--$h_2$ interactions & ??.?? \\
%\hline 
%LEXICAL+syntax25+wiki50 & log. regression (no hidden) & ??.?? \\
%\hline 
{\bf IV}&\multicolumn{2}{l}{\bf Comparison to previous results on WMT12}\\
\midrule
&\disco~\cite{discoMT:WMT2014} & Best on the WMT12 dataset & \emph{na} & \emph{na} & \emph{na} & \emph{na} &  30.5 \\
&\spede & 1st at the WMT12 competition & 21.2 & 27.8 & 26.5 & 26.0 & 25.4\\
&\meteor$^{*}$    & 2nd at WMT12 the competition  & 21.2 & 27.5 & 24.9 & 25.1 & 24.7 \\
&\cite{guzman-EtAl:2014} & Preference kernel approach & 23.1 & 25.8 & 22.6 & 23.2 & 23.7\\
&\amber     & 3rd at the WMT12 competition  & 19.1 & 24.8 & 23.1 & 24.5 & 22.9 \\
\bottomrule 
\end{tabular}}
\caption{\label{t:results}{\small Kendall's tau ($\tau$) on the WMT12 dataset for various metrics.
Notes: \Ni the version of \meteor \ that took part in the WMT12 competition (marked with $^*$ in section IV of the table) is different from the one used in our experiments (section I of the table),
%\Nii \disco \  is scored with a more generous WMT13-version of $\tau$,
\Nii values marked as \emph{na} were not reported by the authors.}}
\end{table*}

The main findings of our experiments are shown in Table~\ref{t:results}.
Section I of Table~\ref{t:results} shows the results for four commonly-used metrics for MT evaluation that compare a translation hypothesis to the reference(s)
using primarily lexical information like word and $n$-gram overlap (even though some allow paraphrases): \bleu, \nist, \ter, and \meteor~\cite{Papineni:Roukos:Ward:Zhu:2002,Doddington:2002:AEM,Snover06astudy,Denkowski2011}.
We will refer to the set of these four metrics as \metrics.
These metrics are not tuned and achieve Kendall's $\tau$ between 18.5 and 23.5.

Section II of Table~\ref{t:results} shows the results for multi-layer neural networks trained on vectors from word embeddings only: \syntax \ and \wiki.
These networks achieve modest $\tau$ values around 10, which should not be surprising: they use very general vector representations and have no access to word or $n$-gram overlap or to length information, which are very important features to compute similarity against the reference. However, as will be discussed below, their contribution is complementary to the four previous evaluation metrics and will lead to significant improvements in combination with them.
%They merely map the two hypotheses and the reference into a vector and then they use these three vectors as input to the NNå.

Section III of Table~\ref{t:results} shows the results for neural networks that combine the four metrics from \metrics \ with \syntax \ and \wiki.
We can see that just combining the four metrics in a flat neural net (i.e., no hidden layer), which is equivalent to a logistic regression,
yields a $\tau$ of 27.06, which is better than the best of the four metrics by 3.5 points absolute,
and also better by over 1.5 points absolute than the best metric that participated at the WMT12 metrics task competition (\spede\ with $\tau=25.4$).
Indeed, \metrics \ is a strong mix that involves not only simple lexical overlap but also approximate matching, paraphrases, edit distance, lengths, etc.
Yet, adding to \metrics \ the embedding vectors yields sizeable further improvements: +1.5 and +2.0 points absolute when adding \syntax \ and \wiki, respectively.
Finally, adding both yields even further improvements close to $\tau$ of 30 (+2.64 $\tau$ points), showing that lexical semantics and syntactic representations are complementary.

Section IV of Table~\ref{t:results} puts these numbers in perspective: it lists the $\tau$ for the top three systems that participated at WMT12,
whose scores ranged between 22.9 and 25.4.

We can see that \metrics \ is much stronger than the winner at WMT12, and thus arguably a baseline hard to improve upon.
While our results are slightly behind those of \disco~\cite{discoMT:WMT2014}, we should note that we only combine four metrics, plus the vectors,
while \disco \ combines over 20 metrics, many of which are costly to compute. 
% Commented this as it is not accurate.
%Moreover, the scores published for \disco \ use the relaxed WMT13 version of $\tau$,
%whose values are a bit inflated compared to the pre-WMT13 version that we use here.

On the other hand, we work in a ranking framework, i.e., we are not interested in producing an absolute score, but in making pairwise decisions only.
Mapping these pairwise decisions into an absolute score is challenging and in our experiments it leads to a slight drop in $\tau$ (results omitted here to save space).

The only other result on WMT12 by authors working with our pairwise framework is our own previous work \cite{guzman-EtAl:2014},
where we used a preference kernel approach to combine syntactic and discourse trees with lexical information;
as we can see, our earlier results are 6 absolute points lower than those we achieve here. Moreover, our NN approach offers advantages over SVMs in terms of computational cost.

Based on these results, we can conclude that word embeddings, whether syntactic or semantic, offer generalizations that efficiently complement very strong metric combinations, and thus should be considered when designing future MT evaluation metrics.
%
% \begin{itemize}
%
%
% \item One big table of results containing all baseline models and NN variations. 
%
% %\item Do we need these results at the level of language pairs as well?
%
% \item Discussion I:
%   \begin{itemize}
%   \item Individual BLEU, NIST, TER and METEOR, and their combination by logistic regression
%   \item Contribution of syntax and sentence-vectors alone
%   \item Contribution on top of the lexical-based measures: incremental and sizeable gains
%   \item Comparison to the WMT12 results and best results in the state of the art (our WMT paper), including the EMNLP-2014 paper 
%   \item Need of the interactions between h1 and h2 representations
%   \item Need of the hidden layer at all (logistic regression on top of the input units ---with vector subtracting)
%   \end{itemize}
%
% \item Further discussion II:
%   \begin{itemize}
%   \item Need of capturing the interactions between h1 and h2?
%   \item Need of the hidden layer at all (logistic regression on top of the input units ---with vector subtracting)?
%   \end{itemize}
%  
% \end{itemize}

\section{Discussion}
\label{sec:discussion}
%-------------------------------------------------------
% Neural Machine Translation Evaluation
% Paper submitted to NAACL-2105
% December 2014
%-------------------------------------------------------
% Section 5 1/2: Results-discussion
%-------------------------------------------------------
\vspace{-3pt}
In this section, we explore how different parts of our framework can be modified to improve its performance, or how it can be extended for further generalization. First, we explore variations of the feature sets from the perspective of both the pairwise features and the embeddings. Then, we analyze the role of the network architecture and of the cost function used for learning.

\subsection{Fine-Grained Pairwise Features} %: Decomposing \bleu}

\begin{table*}[t]
\centering
\vspace{-4mm}
{\small\begin{tabular}{llccccc}
\toprule
 &  & \multicolumn{5}{c}{\bf Kendall's $\tau$} \\\cmidrule(l{2pt}r{2pt}){3-7}
\multicolumn{1}{c}{\bf System} & \multicolumn{1}{c}{\bf Details} & { \bf cz }&{\bf de} & {\bf es}  & {\bf fr} &{\bf AVG}\\
\midrule
\bleu & no learning & 15.88 & 18.56 & 18.57 & 20.83 & 18.46\\
\midrule
\bleucomp & logistic regression & 18.18 & 21.13 & 19.79 & 19.91 & 19.75\\
\bleucomp+\syntax & multi-layer NN & 20.75 & 25.32 & 24.85 & 23.88 & 23.70\\
\bleucomp+\wiki & multi-layer NN & 22.96 & 26.63 & 25.99 & 24.10 & 24.92\\
\bleucomp+\syntax+\wiki & multi-layer NN & 22.84 & 28.92 & 27.95 & 24.90 & \bf 26.15\\
\midrule 
{\it \bleu+\syntax+\wiki} & \it multi-layer NN & \it 20.03 & \it 25.95 & \it 27.07 & \it 23.16 & \it 24.05\\
\bottomrule
\end{tabular}}
\caption{\label{t:results:BLEUcomp}{\small Kendall's $\tau$ on WMT12 for neural networks using \bleucomp, a decomposed version of \bleu. For comparison, the last line shows a combination using \bleu \ instead of \bleucomp.} }
\end{table*}

We have shown that our NN can integrate syntactic and semantic vectors
with scores from other metrics.
In fact, ours is a more general framework,
where one can integrate the \emph{components of a metric} instead of its score,
which could yield better learning.
Below, we demonstrate this for \bleu.

\bleu \
%is the most popular metric for MT evaluation and it
%has components that our vectors lack:
%$n$-gram matches and length ratios.
%On the other hand, our vectors model syntax and semantics beyond exact word matching.
%Thus, it is natural to try to put them together.
%For this purpose, we use as input the components of BLEU:
has different components: 
the $n$-gram precisions,
the $n$-gram matches,
the total number of $n$-grams ($n$=1,2,3,4),
the lengths of the hypotheses and of the reference,
the length ratio between them,
and \bleu's brevity penalty.
We will refer to this decomposed \bleu \ as \bleucomp.
Some of these features were previously used in \simpbleu~\cite{song-cohn:2011:WMT}.

The results of using the components of \bleucomp \ as features are shown in Table~\ref{t:results:BLEUcomp}.
We see that using 
a single-layer neural network,
which is equivalent to logistic regression,
outperforms \bleu\ by more than +1 $\tau$ points absolute.

As before, adding \syntax\ and \wiki\ improves the results, but now by a more sizable margin: +4 for the former and +5 for the latter.
Adding both yields +6.5 improvement over \bleucomp, and almost 8 points over \bleu.

We see once again that the syntactic and semantic word embeddings are complementary to the information sources used by metrics such as \bleu, 
and that our framework can learn from richer pairwise feature sets such as \bleucomp. 
%and that there is a lot of potential in combining the information sources directly:
%individually, the vectors achieve $\tau$ of just 10, and \bleu \ only achieves 18,
%but putting these together yields 26, which is higher by 1 point absolute than the score of the winner at WMT12. Again, the results are also consistent across languages.
%Moreover, the last line of the table shows that using the fine-grained components of \bleu has additive improvements to the \ instead of th \bleucomp \ in the combination yields significant drop in performance, which suggests that it is better to use as input the components of a metric rather than the metric score.

%Finally, the full combination \metrics+ \syntax+\wiki+\bleucomp\ achieves 29.75 (not shown in the table for the sake of simplicity),
%which is slightly better than the 29.70 for \metrics+ \syntax+\wiki.

\subsection{Larger Semantic Vectors}

One interesting aspect to explore is the effect of the dimensionality of the input embeddings. Here, we studied the impact of using semantic vectors of bigger sizes, trained on different and larger text collections. The results are shown in Table~\ref{t:results:larger}.
We can see that, compared to the 50-dimensional \wiki,
300-400 dimensional vectors are generally better by 1-2 $\tau$ points absolute when used in isolation;
however, when used in combination with \metrics+\syntax, they do not offer much gain (up to +0.2),
and  in some cases, we observe a slight drop in performance. We suspect that the variability across the different collections is due to a domain mismatch. Yet, we defer this question for future work.

\begin{table}[t]
\centering
{\small\begin{tabular}{lcc}
\toprule
{ \bf Source }& { \bf Alone}  & { \bf Comb.}\\
\midrule
\wiki         & \it 10.01 & \it 29.70\\
\wikii        &  9.66 & \bf 29.90\\
\commoncrawl  & \bf 12.16 & 29.68\\
\commoncrawll & \bf 11.41 & \bf 29.88\\
\midrule
\wordvec      &  7.72 & 29.13\\
\midrule
\composes     & \bf 12.35 & 28.54\\
\bottomrule
%\metrics+\syntax+\wiki & multi-layer NN & 26.30 & 33.19 & 30.38& 28.92 & 29.70 \\
%\metrics+\syntax+\wikii & multi-layer NN & 26.30 & 34.13 & 30.07 & 29.11 & \bf 29.90\\
%\metrics+\syntax+\commoncrawl & multi-layer NN & 25.91 & 33.32 & 30.22 & 29.27 & 29.68\\
%\metrics+\syntax+\commoncrawll & multi-layer NN & 25.80 & 33.11 & 31.50 & 29.13 & 29.88\\
\end{tabular}}
\caption{\label{t:results:larger}{\small Average Kendall's $\tau$ on WMT12 for semantic vectors
trained on different text collections.
Shown are results \Ni when using the semantic vectors alone,
and \Nii when combining them with \metrics \ and \syntax.
The improvements over \wiki \ are marked in bold.}}
\end{table}

%!TEX root = neural-mte.tex

%-------------------------------------------------------
% Neural Machine Translation Evaluation
% Paper submitted to NAACL-2105
% December 2014
%-------------------------------------------------------
% Section 6: Discussion
%-------------------------------------------------------

\subsection{Deep vs. Flat Neural Network}

One interesting question is how much of the learning is due to the rich input representations,
and how much happens because of the architecture of the neural network.
To answer this, we experimented with two settings: a single-layer neural network, where all input features are fed directly to the output layer (which is logistic regression), and our proposed multi-layer neural network.

The results are shown in Table~\ref{t:architectures}.
We can see that switching from our multi-layer architecture to a single-layer one yields an absolute drop of 0.6 $\tau$. This suggests that there is value in using the deeper, pairwise layer architecture.

\begin{table}[tb]
\centering
\small\begin{tabular}{lccccc}
\toprule
 & \multicolumn{5}{c}{\bf Kendall's $\tau$} \\\cmidrule(l{2pt}r{2pt}){2-6}
\multicolumn{1}{c}{\bf Details} & { \bf cz }&{\bf de} & {\bf es}  & {\bf fr} &{\bf AVG}\\
\midrule
 single-layer & 25.86 & 32.06 & 30.03& 28.45 &  29.10 \\
 multi-layer  & 26.30 & 33.19 & 30.38& 28.92 & {\bf 29.70} \\
\bottomrule
\end{tabular}
\caption{{\small\label{t:architectures} Kendall's tau ($\tau$) on the WMT12 dataset for alternative architectures using \metrics+\syntax+\wiki ~as input.} }
\vspace{-0.25cm}
\end{table}

% \begin{table*}[tbh]
% \centering
% \small\begin{tabular}{|l|l|ccccc|}
% \hline
%  &  & \multicolumn{5}{|c|}{Kendall's $\tau$} \\
% \multicolumn{1}{|c|}{System} & \multicolumn{1}{|c|}{Details} & cz & de & es & fr & AVG\\
% \hline
% \metrics+\syntax+\wiki & single-layer NN & 25.86 & 32.06 & 30.03& 28.45 &  29.10 \\
% \metrics+\syntax+\wiki & multi-layer NN & 26.30 & 33.19 & 30.38& 28.92 & {\bf 29.70} \\
% \hline
% \end{tabular}
% \caption{\label{t:architectures}Kendall's tau ($\tau$) on the WMT12 dataset for altenrative architectures.}
% \end{table*}

\subsection{Task-Specific Cost Function}

Another question is whether the log-likelihood cost function $J(\theta)$ (see Section~\ref{logcost}) is the most appropriate for our ranking task, provided that it is evaluated using Kendall's $\tau$ as defined below:
\begin{equation}
\tau = \frac{concord. - disc. - ties}{concord+disc. + ties}
\end{equation}

\noindent where \emph{concord.}, \emph{disc.} and \emph{ties} are the number of concordant, disconcordant and tied pairs. 

Given an input tuple $(t_1,t_2,r)$, the logistic cost function yields larger values of $\sigma = f(t_1,t_2,r)$ if $y=1$, and smaller if $y=0$,
where $0 \le \sigma \le 1$ is the parameter of the Bernoulli distribution. However, it does not model \emph{directly} the probability when the order of the hypotheses in the tuple is reversed, i.e., $\sigma'=f(t_2,t_1,r)$. 

For our specific task, given an input tuple $(t_1,t_2,r)$, we want to make sure that the difference between the two output activations $\Delta = \sigma - \sigma'$ is positive when $y=1$, and negative when $y=0$. %.  increase the probability of making a mistake, i.e. the difference between $\sigma_{12}$ and $\sigma_{21}=f(t_2,t_1,r)$ . We may want to promote $\sigma_{12}$ more towards $1$ or $0$ depending on $\sigma_{21}$ and $y$
Ensuring this would take us closer to the actual objective, which is Kendall's $\tau$. One possible way to do this is to introduce a task-specific cost function that penalizes the disagreements
similarly to the way Kendall's $\tau$ does.\footnote{Other variations for ranking tasks are possible, e.g.,~\cite{yih-EtAl:2011:CoNLL}.} %$\Delta = \sigma_{12} - \sigma_{21}$ and let 
 In particular, we define a new \emph{Kendall cost} as follows:
%, to ensure that $\sigma_{12} > \sigma_{21}$ by a larger margin, 
\begin{equation}
J_{\mathbf{\theta}} = - \sum_{n} y_n\sig(-\gamma \Delta_n) + (1-y_n)\sig (\gamma \Delta_n)\,
\end{equation}

\noindent where we use the sigmoid function $\sig$ as a differentiable approximation to the step function.

The above cost function penalizes disconcordances, i.e., cases where \Ni $y=1$ but $\Delta <0$, or \Nii when $y=0$ but $\Delta>0$.
However, we also need to make sure that we discourage \emph{ties}.
We do so by adding a zero-mean Gaussian regularization term $\exp(-\beta\Delta^2/2)$ that penalizes the value of $\Delta$ getting close to zero.
Note that the specific values for $\gamma$ and $\beta$ are not really important, as long as they are large.
In particular, in our experiments, we used $\gamma = \beta = 100$. 

\begin{table}[t]
\centering
\vspace{-4mm}
\small\begin{tabular}{lccccc}
\toprule
& \multicolumn{5}{c}{\bf Kendall's $\tau$} \\\cmidrule(l{2pt}r{2pt}){2-6}
 \multicolumn{1}{c}{\bf Details} & { \bf cz }&{\bf de} & {\bf es}  & {\bf fr} &{\bf AVG}\\
%\hline
%\multicolumn{7}{l}{\bf Alternative cost function}\\
\midrule
 Logistic & 26.30 & 33.19 & 30.38& 28.92 & 29.70 \\
 Kendall  & 27.04 & 33.60 & 29.48& 28.54 & 29.53 \\
 Log.+Ken.  & 26.90 & 33.17& 30.40& 29.21 & {\bf 29.92}\\
\bottomrule
\end{tabular}
\caption{{\small\label{t:costfun} Kendall's tau ($\tau$) on  WMT12 for alternative cost functions using \metrics+\syntax+\wiki.} }
\vspace{-0.35cm}
\end{table}

% \begin{table*}[tbh]
% \centering
% \small\begin{tabular}{|l|l|ccccc|}
% \hline
%  &  & \multicolumn{5}{|c|}{Kendall's $\tau$} \\
% \multicolumn{1}{|c|}{System} & \multicolumn{1}{|c|}{Details} & cz & de & es & fr & AVG\\
% %\hline
% %\multicolumn{7}{l}{\bf Alternative cost function}\\
% \hline
% \metrics+\syntax+\wiki & Logistic cost& 26.30 & 33.19 & 30.38& 28.92 & 29.70 \\
% \metrics+\syntax+\wiki & Kendall cost & 27.04 & 33.60 & 29.48& 28.54 & 29.53 \\
% \metrics+\syntax+\wiki & Logistic+Kendall cost & 26.90 & 33.17& 30.40& 29.21 & {\bf 29.92}\\
% \hline 
% \end{tabular}
% \caption{\label{t:costfun}Kendall's tau ($\tau$) on the WMT12 dataset for alternative cost functions.}
% \end{table*}

Table~\ref{t:costfun} shows a comparison of the two cost functions: \Ni the standard logistic cost, and \Nii our Kendall cost.
We can see that using the Kendall cost enables effective learning, although it is eventually outperformed by the logistic cost.
Our investigation revealed that this was due to a combination of slower convergence and poor initialization.
Therefore, we further experimented with a setup where we first used the logistic cost to pre-train the neural network,
and then we switched to the Kendall cost in order to perform some finer tuning.
As we can see in Table~\ref{t:costfun} (last row), doing so yielded a sizable improvement over using the Kendall cost only;
it also improved over using the logistic cost only.
 
%Additionally, we include the results of using the 
%

%  However, when we use the Kendall cost to fine-tune models trained by the log-likelihood function, we achieve a sizable improvement in Kendall's tau.
%\end{itemize}

\section{Conclusions and Future Work}
\label{sec:conclusions}

%-------------------------------------------------------
% Neural Machine Translation Evaluation
% Paper submitted to NAACL-2105
% December 2014
%-------------------------------------------------------
% Section 8: Conclusions (and further work)
%-------------------------------------------------------

%We have presented a novel neural-based architecture for MT evaluation,
%where lexical, syntactic and semantic information from the reference and the two hypotheses is %compacted into relatively small distributed vector representations
%%(which are in turn learned using deep learning)
%and fed into the NN's input layer. 

We have presented a novel framework for learning a tunable MT evaluation metric
in a pairwise ranking setting,
given pre-existing pairwise human preference judgments.

In particular, we used a neural network,
where the input layer encodes lexical, syntactic and semantic information from the reference and the two translation 
hypotheses, which is efficiently compacted into relatively small embeddings. %distributed vector representations
%(which are in turn learned using deep learning)
%and fed into the NN's input layer. 
The network has a hidden layer, motivated by our intuition about the problem, which captures the interactions 
between the relevant input components.
Unlike previously proposed kernel-based approaches, our framework allows us to do both training and inference efficiently.
Moreover, we have shown that it can be trained to optimize a task-specific cost function,
which is more appropriate for the pairwise MT evaluation setting.

%A hidden layer, motivated by our intuitions on the pairwise ranking problem, is used to capture interactions between the relevant input components. Finally, we use a task oriented cost function, specifically tailored for this problem.

%that models complex pairwise-relations between two translation and a reference, using lexical, semantic and syntactic features. Our framework is designed to allow the incorporation of neural embeddings, along with other useful sources of information. Furthermore, it can successfully be trained to optimize task-specific cost functions more appropriate for the MT evaluation setting.

The evaluation results have shown that our NN model yields state-of-the-art results when using lexical, syntactic and semantic features
(the latter two based on compact embeddings).
Moreover, we have shown that the contribution of the different information sources is additive, thus demonstrating that 
the framework can effectively integrate complementary information.
Furthermore, the framework is flexible enough to exploit different granularities of features such as $n$-gram matches and other components of \bleu \ (which individually work better than using the aggregated \bleu \ score).
%thus our model performs better the more information is presented with. 
Finally, we have presented evidence suggesting that using the pairwise hidden layers is advantageous over simpler flat models.

%The contribution of the paper is twofold: i) using distributed representations of syntax and lexical semantics for MT evaluation in a flexible combination framework; ii) training a NN architecture with a hidden layer to capture interesting pairwise interactions. We don't consider the distributed representations of syntax and semantics a minor aspect, as they capture information not present in a strong combination of individual lexical evaluation metrics (BLEU+NIST+METEOR+TER). Kendall's Tau goes from 27.06 to 29.70, a sizeable improvement according to the state-of-the-art. 

In future work, we would like to experiment with an extension that allows for multiple references.
We further plan to incorporate features from the \emph{source} sentence. We believe that our framework can support learning similarities between the two translations and the source, for an improved MT evaluation. Variations of this architecture might be useful for related tasks such as Quality Estimation and hypothesis re-ranking for Machine Translation, where no references are available.

Other aspects worth studying as a complement to the present work include \Ni the impact of the quality of the syntactic analysis (translations are often just a ``word salad''), \Nii differences across language pairs, and \Niii the relevance of the domain the semantic representations are trained on.

%%% For future work
%Reviewer's comments on i)the quality of the input representations, ii)the observed language-pair differences, and iii)the domain relevance are all very interesting and deserve further research. However, they are beyond the scope of the paper. Our position in the current version is summarized below:

%-The underlying assumption regarding the generation of the distributed representations is not that tools like the Stanford Parser are error free, but rather that they are consistent. The MT output is undoubtedly noisy and the parser has difficulties in analyzing some of the ungrammatical sentences, but we think that the neural framework is robust enough to learn to use these differences. 

%-The overall improvements when adding different layers of information are observed in all language pairs (admittedly, with different absolute differences), which adds on the reliability of the approach. 
  
%-More than using domain-specific embedding models, our current approach relies on using general models trained on large text collections representing a variety of domains and genres. 

%----------------------- A c k n o w l e d g e m e n t s
%\section*{Acknowledgments}

%----------------------- B i b l i o g r a  p h y
\bibliographystyle{acl}
\bibliography{bibliography/ACL} % this started as a copy of EMNLP bibliography style

\end{document}